\documentclass[10pt, a4paper]{article}

\usepackage[final]{acl} 
\usepackage{amssymb}    
\usepackage{amsthm}
\usepackage{xcolor}
\usepackage{booktabs}
\usepackage{tikz}
\usetikzlibrary{positioning}
\usepackage{listings}
\usepackage{xcolor}
\usepackage{booktabs}
\usepackage{tabularx}
\usepackage{amsmath}
\usepackage{times}
\usepackage{latexsym}
\usepackage[T1]{fontenc}
\usepackage[utf8]{inputenc}
\usepackage{microtype}
\usepackage{inconsolata}
\usepackage{graphicx}

\definecolor{jsonkey}{HTML}{1C00CF}
\definecolor{jsonstring}{HTML}{C41A16}
\definecolor{jsonnumber}{HTML}{1A7F37}
\definecolor{jsonbackground}{HTML}{F9F9F9}
\definecolor{jsongray}{gray}{0.3}

\lstdefinelanguage{json}{
  basicstyle=\ttfamily\scriptsize,
  backgroundcolor=\color{jsonbackground},
  showstringspaces=false,
  breaklines=true,
  frame=single,
  framerule=0.3pt,
  rulecolor=\color{gray!40},
  stringstyle=\color{jsonstring},
  numberstyle=\color{jsonnumber},
  literate=
   *{:}{{{\color{jsongray}{:}}}}{1}%
    {,}{{{\color{jsongray}{,}}}}{1}%
    {"}{{{\color{gray}{\texttt{"}}}}}{1}%
    {"}{{{\color{gray}{\texttt{"}}}}}{1},
  morestring=[b]",
  commentstyle=\color{gray},
  keywordstyle=\color{jsonkey}\bfseries,
}

\title{Timing In stand-up Comedy: Text, Audio, Laughter, Kinesics (TIC-TALK): \\ Pipeline and Database for the Multimodal Study of Comedic Timing}

\author{
Yaelle Zribi\textsuperscript{1} \quad
Florian Cafiero\textsuperscript{1,2} \quad
Vincent L\'epinay\textsuperscript{3} \quad
Chahan Vidal-Gor\`ene\textsuperscript{1} \\[0.5em]
\textsuperscript{1}Centre Jean-Mabillon, \'Ecole nationale des chartes, PSL, Paris, France \\
\textsuperscript{2}Laboratoire de Recherche, EPITA, Paris, France \\
\textsuperscript{3}m\'edialab, Sciences Po, Paris, France \\
\texttt{\{yaelle.zribi, florian.cafiero, chahan.vidal-gorene\}@chartes.psl.eu} \\
\texttt{vincent.lepinay@sciencespo.fr}
}

\begin{document}

\maketitle

\begin{abstract}
Stand-up comedy, and humor in general, are often studied through their verbal content. Yet live performance relies just as much on embodied presence and audience feedback. We introduce TIC-TALK, a multimodal resource with 5,400+ temporally aligned topic segments capturing language, gesture, and audience response across 90 professionally filmed stand-up comedy specials (2015–2024). The pipeline combines BERTopic for 60\,s thematic segmentation with dense sentence embeddings, Whisper-AT for 0.8\,s laughter detection, a fine-tuned YOLOv8-cls shot classifier, and YOLOv8s-pose for raw keypoint extraction at 1\,fps. Raw 17-joint skeletal coordinates are retained without prior clustering, enabling the computation of continuous kinematic signals—arm spread, kinetic energy, and trunk lean—that serve as proxies for performance dynamics. All streams are aligned by hierarchical temporal containment without resampling, and each topic segment stores its sentence-BERT embedding for downstream similarity and clustering tasks. As a concrete use case, we study laughter dynamics across 24 thematic topics: kinetic energy negatively predicts audience laughter rate ($r=-0.75$, $N=24$), consistent with a stillness-before-punchline pattern; personal and bodily content elicits more laughter than geopolitical themes; and shot close-up proportion correlates positively with laughter ($r=+0.28$), consistent with reactive montage.
\end{abstract}

\section{Introduction}

Humor is one of the most complex forms of human communication, involving timing, embodiment, and interaction. Stand-up comedy, in particular, constitutes an ideal case study for modeling how verbal, acoustic, and visual cues align to produce shared affective meaning.

The performer’s gestures, pauses, and interaction with audience laughter are essential components of meaning and rhythm. Modeling these elements jointly raises both technical and conceptual challenges: how can we represent and evaluate \textit{comedic timing} computationally? We take \textit{comedic timing} to denote short-lag coordination across text, gesture, and audience response in live delivery.

Multimodal modeling of live performance is still mostly missing. Computational humor has largely centered on \emph{textual} humor and its assessment \cite{kalloniatis-adamidis-2025-review}: detecting humorous passages \cite{weller2019humor, annamoradnejad2024colbert, yang-etal-2015-humor, cafiero2025riddle, hossain-etal-2020-semeval}, ranking by funniness \cite{potash-etal-2017-semeval}, and predicting offensiveness or controversy \cite{meaney-etal-2021-semeval}.

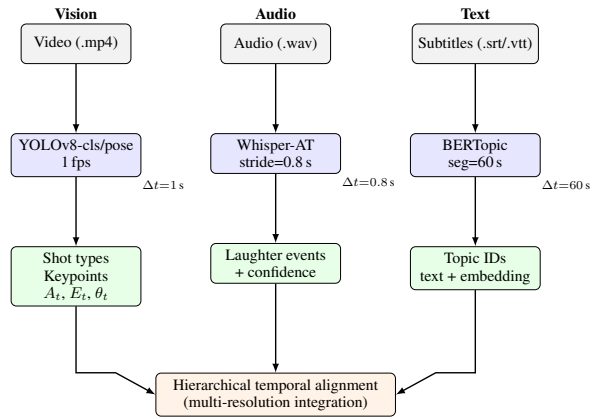
\begin{figure}[ht]
\centering
\resizebox{\columnwidth}{!}{%
\begin{tikzpicture}[
    font=\small,
    node distance=1.4cm and 1.8cm,
    every node/.style={align=center},
    data/.style={draw, rounded corners=3pt, fill=gray!10, minimum width=2.2cm, minimum height=0.8cm},
    process/.style={draw, rounded corners=3pt, fill=blue!10, minimum width=2.6cm, minimum height=0.8cm},
    result/.style={draw, rounded corners=3pt, fill=green!10, minimum width=2.6cm, minimum height=0.8cm},
    merge/.style={draw, rounded corners=3pt, fill=orange!10, minimum width=4.8cm, minimum height=0.8cm},
    arrow/.style={->, thick, >=latex}
]

\node at (0,5.6) {\textbf{Vision}};
\node at (4,5.6) {\textbf{Audio}};
\node at (8,5.6) {\textbf{Text}};

\node[data] (video) at (0,5) {Video (.mp4)};
\node[data] (audio) at (4,5) {Audio (.wav)};
\node[data] (sub)   at (8,5) {Subtitles (.srt/.vtt)};

\node[process, below=of video] (vision) {YOLOv8-cls/pose \\ 1\,fps};
\node[process, below=of audio] (sound) {Whisper-AT \\ stride=0.8\,s};
\node[process, below=of sub]   (text)  {BERTopic \\ seg=60\,s};

\node[result, below=of vision] (visionout) {Shot types \\ Keypoints \\ $A_t$, $E_t$, $\theta_t$};
\node[result, below=of sound]  (soundout)  {Laughter events \\ + confidence};
\node[result, below=of text]   (textout)   {Topic IDs \\ text + embedding};

\node[merge, below=1.8cm of soundout] (merge)
  {Hierarchical temporal alignment \\ (multi-resolution integration)};

\draw[arrow] (video) -- (vision);
\draw[arrow] (audio) -- (sound);
\draw[arrow] (sub)   -- (text);

\draw[arrow] (vision) -- (visionout);
\draw[arrow] (sound)  -- (soundout);
\draw[arrow] (text)   -- (textout);

\draw[arrow] (visionout) |- ([xshift=-0.8cm]merge.north west) -- (merge.west);
\draw[arrow] (soundout)  -- (merge);
\draw[arrow] (textout)   |- ([xshift=0.8cm]merge.north east) -- (merge.east);

\node[below right=-0.05cm and -0.1cm of vision] {\scriptsize $\Delta t{=}1$\,s};
\node[below right=-0.05cm and -0.1cm of sound]  {\scriptsize $\Delta t{=}0.8$\,s};
\node[below right=-0.05cm and -0.1cm of text]   {\scriptsize $\Delta t{=}60$\,s};

\end{tikzpicture}%
}
\caption{Multimodal processing pipeline. Each modality retains its native temporal resolution. Signals are merged by hierarchical temporal containment without resampling. Topic segments also store 384-dim sentence-BERT embeddings (\texttt{all-MiniLM-L6-v2}).}
\label{fig:pipeline}
\end{figure}

Distant viewing offers a framework to tackle the conceptual and technical challenge to simulate human vision for artistic analysis \cite{arnold2019distant}. Affective and social computing provide tools for modeling prosody, gesture, and emotion, though primarily in human–machine interaction.

Movement analysis has been applied to iconography \cite{impett2017totentanz}, ergonomic gesture learning \cite{glushkova2023glass}, and pose clustering in archival theater footage \cite{rau2023ai} but without linking gesture to speech or audience response.

Recent work explores how language models might support stand-up writing, raising questions of prompt design, model bias, and cultural framing \cite{mirowski-etal-2024-robot-bar}. The StandUp4AI dataset \cite{barriere2025standup4ai} adds multilingual laughter labels, but focuses on audio.

Beyond text-based generation, live performance has been identified as a critical setting for evaluating computational humor under real-world temporal and social constraints. \citet{mirowski-etal-2025-theater} argue that improvised and staged comedy offer unique testbeds for studying the interaction between human performers, AI systems, and audiences, emphasizing that timing, embodiment, and audience feedback are inseparable from humor evaluation. While our focus differs from these experimental setups, we contribute a reproducible multimodal dataset of stand-up performances, designed to operationalize these performative dimensions (timing, gesture, and laughter) as measurable signals for computational analysis.

Recent work has similarly proposed an automatic method for the analysis of stand-up comedy, with particular attention to timing and performance dynamics. Using repeated recordings of the same routines performed on different nights by two comedians, \citet{pope2026timing} analyze timing structures in live comedy through matched speech sequences across performances, demonstrating that the apparent spontaneity of stand-up relies on a complex craft of pacing and adaptation to audience context. Their study shares our interest in timing, performance structure, and audience response, but differs in both scale and design: it focuses on prosody, speech, and laughter placement in unedited live performances, whereas we analyze multimodal coordination within a large corpus of professionally recorded and edited stand-up comedy specials, examining themes, laughter placement, as well as the visual and bodily dimensions of performance through shot composition and pose-derived movement signals. Taken together, these contrasting approaches underscore how central yet elusive timing remains in the analysis of stand-up comedy.

In this paper, we present TIC-TALK: a multimodal corpus of 90 Netflix stand-up specials and a documented processing pipeline aligning text, audio, and vision without resampling. We report available performance indicators for the shot classifier and descriptive coverage metrics for the other streams, and include corpus-level cross-modal findings—notably a negative relationship between kinetic energy and laughter rate across topics—that validate the temporal alignment.

Our main contributions are:
\begin{enumerate}
    \item A reusable multimodal \textbf{corpus} of 90 professionally edited stand-up specials with temporally aligned text, audio, and vision streams, including raw pose keypoints and per-segment sentence-BERT embeddings;
    \item A transparent, reproducible \textbf{processing pipeline} (BERTopic, sentence-BERT, Whisper-AT, YOLOv8-cls, YOLOv8s-pose) with documented training choices and performance indicators;
    \item \textbf{Three kinematic signals} ($A_t$, $E_t$, $\theta_t$) derived from raw skeletal coordinates; and a \textbf{cross-modal use case} (Section~\ref{sec:crossmodal}) on laughter dynamics across 24 thematic topics, demonstrating: (i) kinetic energy is the strongest kinematic predictor of laughter ($r=-0.75$, $N=24$), consistent with a stillness-before-punchline pattern; (ii) personal/bodily themes elicit systematically more laughter than geopolitical content; (iii) belly laughs are virtually absent at topic granularity, motivating event-level annotation; and (iv) shot close-up proportion correlates weakly with laughter rate ($r=+0.28$), consistent with reactive montage;
    \item A \textbf{short-horizon laughter onset prediction benchmark} (Section~\ref{sec:onset}): given multimodal context up to $t$, predict whether a new laughter event will begin in $[t, t{+}2)$; ablation over five feature sets ($N{=}285{,}916$ anchors, 90 shows, show-level train/val/test split) shows that temporal laughter history dominates prediction (AUROC\,=\,0.643), that multimodal fusion achieves the best performance (AUROC\,=\,0.647, AUPRC\,=\,0.277 vs.~random baseline 0.170), and that shot and pose features contribute marginal but consistent gains.
\end{enumerate}

We first detail the processing pipeline per modality and the temporal alignment strategy (Section~\ref{sec:alignment}), then describe the corpus as a reusable resource and its data structure (Section~\ref{sec:corpus}). Section~\ref{sec:crossmodal} presents a descriptive use case---laughter dynamics across thematic topics; Section~\ref{sec:onset} presents a predictive use case---short-horizon laughter onset prediction. Limitations and biases are discussed in Section~\ref{sec:discussion}.

\section{Processing Pipeline and Temporal Alignment}
\label{sec:alignment}

The dataset integrates four modality streams—text, audio, and vision (pose and shot)—each processed independently at its native temporal resolution before alignment into a unified hierarchical representation (Figure~\ref{fig:pipeline}).

\subsection{Audio: Laughter Detection}

Laughter events were detected using \textbf{Whisper-AT}, a pretrained AudioSet-based audio tagging model~\cite{Gong2023WhisperAT}.
Inference was performed at a 0.8\,s stride in a high-recall configuration. The model outputs class probabilities for multiple laughter types; contiguous positive windows were merged into continuous events.  
Each event is represented by start and end times in seconds, a label (\texttt{type}), and a confidence score. These events constitute high-resolution temporal anchors for audience response.

\subsection{Text: Topic Segmentation}

\subsubsection{Data and time-based segmentation}
We start from subtitle transcripts in \texttt{.srt} format, parsed with their start/end timestamps and normalized by removing markup and formatting codes, standardizing apostrophes, and collapsing whitespace. We then construct contiguous time blocks by concatenating consecutive subtitle lines until a target duration is reached; a new block starts at the next subtitle line whose start time exceeds the current block limit. We remove stopwords using the union of a standard English stopword list and a curated set targeting fillers/discourse markers.

\subsubsection{Sentence embeddings}
Each block is embedded with a sentence-transformer encoder (\texttt{all-MiniLM-L6-v2}). Embeddings are computed in batches and L2-normalized: for each block $i$ we obtain an embedding $\mathbf{e}_i \in \mathbb{R}^d$ and set
\[
\tilde{\mathbf{e}}_i = \frac{\mathbf{e}_i}{\|\mathbf{e}_i\|_2}.
\]
These normalized vectors are used both for training the topic model and for later topic assignment to 60-second blocks.

\subsubsection{Topic modeling with BERTopic}
We learn topics using BERTopic. We first apply UMAP to the normalized embeddings using cosine distance, with $n_{\text{neighbors}}=15$, $n_{\text{components}}=5$, and $min\_dist=0.1$. The reduced representations are clustered with HDBSCAN ($min\_cluster\_size=15$, $min\_samples=5$). Topic representations are obtained from a unigram--bigram count vectorizer. The model is capped at $40$ topics and we retain the top $10$ words per topic for interpretation.

\subsubsection{Model selection over training block size}
\label{sec:model-selection}
Topic quality depends on the training granularity. We select the training block size from $\{120,150,180,210,240\}$ seconds. For each candidate size, we train a BERTopic model and compute three diagnostics based on the resulting topic assignments:
\begin{itemize}
    \item the \emph{number of discovered topics} $K$, 
    \item  the \emph{largest-topic share} $s_{\max}$
    \item  a \emph{normalized topic entropy} $H_{\text{norm}}$.
\end{itemize}

Let $n_k$ be the number of blocks assigned to topic $k \in \{1,\dots,K\}$ and $N=\sum_{k=1}^K n_k$. Define $p_k = n_k / N$. We compute
\[
s_{\max} = \max_{k \in \{1,\dots,K\}} p_k \] and
\[H_{\text{norm}} = \frac{-\sum_{k=1}^K p_k \log p_k}{\log K}.
\]
We enforce two validity constraints to avoid degenerate solutions: $K \ge 10$ and $s_{\max} \le 0.35$. Among valid candidates, we select the model maximizing a composite score:
\[
S = H_{\text{norm}} + C_{\text{NPMI}} - 2\, s_{\max},
\]
where $C_{\text{NPMI}}$ is a topic coherence measure computed from the model's top words.

\paragraph{Topic coherence (NPMI).}
We compute coherence using normalized pointwise mutual information (NPMI) over a tokenized version of the preprocessed blocks (subsampling documents when necessary for efficiency). For a topic $t$, let $W_t$ be its top-$M$ words (here $M=10$). For each pair $(w_i,w_j)\in W_t \times W_t$ with $i<j$, let $P(w)$ be the probability that a word appears in a document and $P(w_i,w_j)$ the probability that both appear in the same document (estimated from document co-occurrence counts). NPMI for a pair is
\[
\operatorname{NPMI}(w_i,w_j)
=
\frac{\log \frac{P(w_i,w_j)}{P(w_i)P(w_j)}}{-\log P(w_i,w_j)}.
\]
Topic coherence is the average over word pairs, and corpus-level coherence averages over topics:
\[
C_{\text{NPMI}}
=
\frac{1}{|\mathcal{T}|}\sum_{t\in\mathcal{T}}
\left(
\frac{2}{M(M-1)} \sum_{i<j} \operatorname{NPMI}(w_i,w_j)
\right),
\]
where $\mathcal{T}$ is the set of non-outlier topics.

\subsubsection{Topic assignment to 60-second segments}
\label{sec:assignment}
After selecting the training block size, we retrain the topic model on all blocks at that granularity. We then construct 60-second blocks for each show and assign a topic label to each block using the trained BERTopic model. Since HDBSCAN can label uncertain points as outliers ($-1$), we apply a three-step post-processing procedure to reduce outliers
(1) BERTopic outlier reduction using an embedding-based strategy followed by a c-TF--IDF-based strategy,
(2) reassignment of remaining outliers by nearest topic centroid in embedding space, and
(3) within-show temporal gap filling: if a single outlier is flanked by two identical non-outlier topics, it is replaced by that topic.

For the centroid step, we compute a centroid for each non-outlier topic $k$ by averaging the normalized training embeddings assigned to $k$ and re-normalizing:
\[
\mathbf{c}_k = \frac{\frac{1}{|I_k|}\sum_{i\in I_k}\tilde{\mathbf{e}}_i}{\left\|\frac{1}{|I_k|}\sum_{i\in I_k}\tilde{\mathbf{e}}_i\right\|_2},
\]
where $I_k$ is the set of training blocks assigned to topic $k$. For an outlier segment with embedding $\tilde{\mathbf{e}}$, we compute cosine similarities $\tilde{\mathbf{e}}^\top \mathbf{c}_k$ and reassign it to $\arg\max_k \tilde{\mathbf{e}}^\top \mathbf{c}_k$ when the maximum similarity is at least $0.30$.

This procedure yields a topic sequence at 60-second resolution for each show, together with interpretable topic descriptors derived from c-TF--IDF.

\subsection{Vision: Model Training and Feature Extraction}

The visual pipeline combines two complementary YOLOv8 models operating hierarchically:  
(1) a \textbf{YOLOv8-cls} network fine-tuned for shot classification, and  
(2) a pre-trained \textbf{YOLOv8s-pose} estimator used to extract body keypoints from full-body frames only~\cite{maji2022yolo}.
This design ensures both efficient processing and consistent framing for pose analysis.

\paragraph{Shot classification.}
Fine-tuned to recognize six shot types:  
\emph{full shot, medium close-up, medium long shot, medium shot, other angles,} and \emph{other}.  
The model was trained on 594 manually annotated frames and validated on 128 held-out samples (100 epochs, batch size 32, learning rate $1\times10^{-3}$). 
Validation yielded an average F1 = 0.91, with most confusion between adjacent framings (e.g., chest $\leftrightarrow$ waist).  
Predicted shot labels later serve both as contextual information and as filters for pose extraction, keeping only full-body and frontal views.

\paragraph{Pose estimation.}
Filtered frames are processed by the pre-trained \textbf{YOLOv8s-pose} model, producing 17 body keypoints (COCO skeleton) per detected performer at 1\,fps.
Raw pixel-normalized coordinates for all 17 joints are stored without prior discretization, preserving the full geometric information for downstream analysis. Joints with no detection confidence are recorded as $(0,0)$ and excluded from derived computations via a validity filter.

\paragraph{Kinematic signals derived from raw keypoints.}
\label{par:kinematics}
Raw keypoints enable the computation of three continuous scalar signals per frame, each capturing a distinct dimension of performance dynamics.
Let $\mathbf{p}_j(t)$ denote the 2D coordinates of joint $j$ at time $t$, and let $J(t)$ be the set of valid joints at $t$.

\textbf{Arm spread} measures the lateral extension of the performer's gesture relative to shoulder width:
\[
A_t = \frac{\|\mathbf{p}_{\text{W}_1}(t) - \mathbf{p}_{\text{W}_2}(t)\|}{\|\mathbf{p}_{\text{S}_1}(t) - \mathbf{p}_{\text{S}_2}(t)\|},
\]
where $\text{W}_1,\text{W}_2$ are the wrists and $\text{S}_1,\text{S}_2$ the shoulders. $A_t{=}1$ corresponds to a neutral stance; $A_t{>}2$ indicates open or emphatic gestures. 

\textbf{Kinetic energy} quantifies total body movement between consecutive frames, normalized by performer height (bounding-box height $h$):
\[
E_t = \frac{1}{h} \sum_{j \in J(t)\cap J(t{-}1)} \|\mathbf{p}_j(t) - \mathbf{p}_j(t{-}1)\|.
\]
This serves as a proxy for performance intensity, capturing transitions between high-agitation delivery and still, high-confidence pauses.

\textbf{Trunk lean} encodes the signed angle of the torso axis relative to vertical:
\[
\theta_t = \arctan\!\left(\frac{x_{\text{hip}}(t) - x_{\text{sho}}(t)}{y_{\text{hip}}(t) - y_{\text{sho}}(t)}\right) \times \frac{180}{\pi},
\]
where $x_{\text{sho}},y_{\text{sho}}$ and $x_{\text{hip}},y_{\text{hip}}$ are the midpoints of the shoulder and hip pairs respectively. Lateral leans may, for instance, be characteristic of character-mimicry and asides, providing a posture-level complement to motion energy.

All three signals are smoothed with a sliding window of 30\,s to suppress frame-level noise before analysis.

\begin{figure*}[t]
\centering
\includegraphics[width=\textwidth]{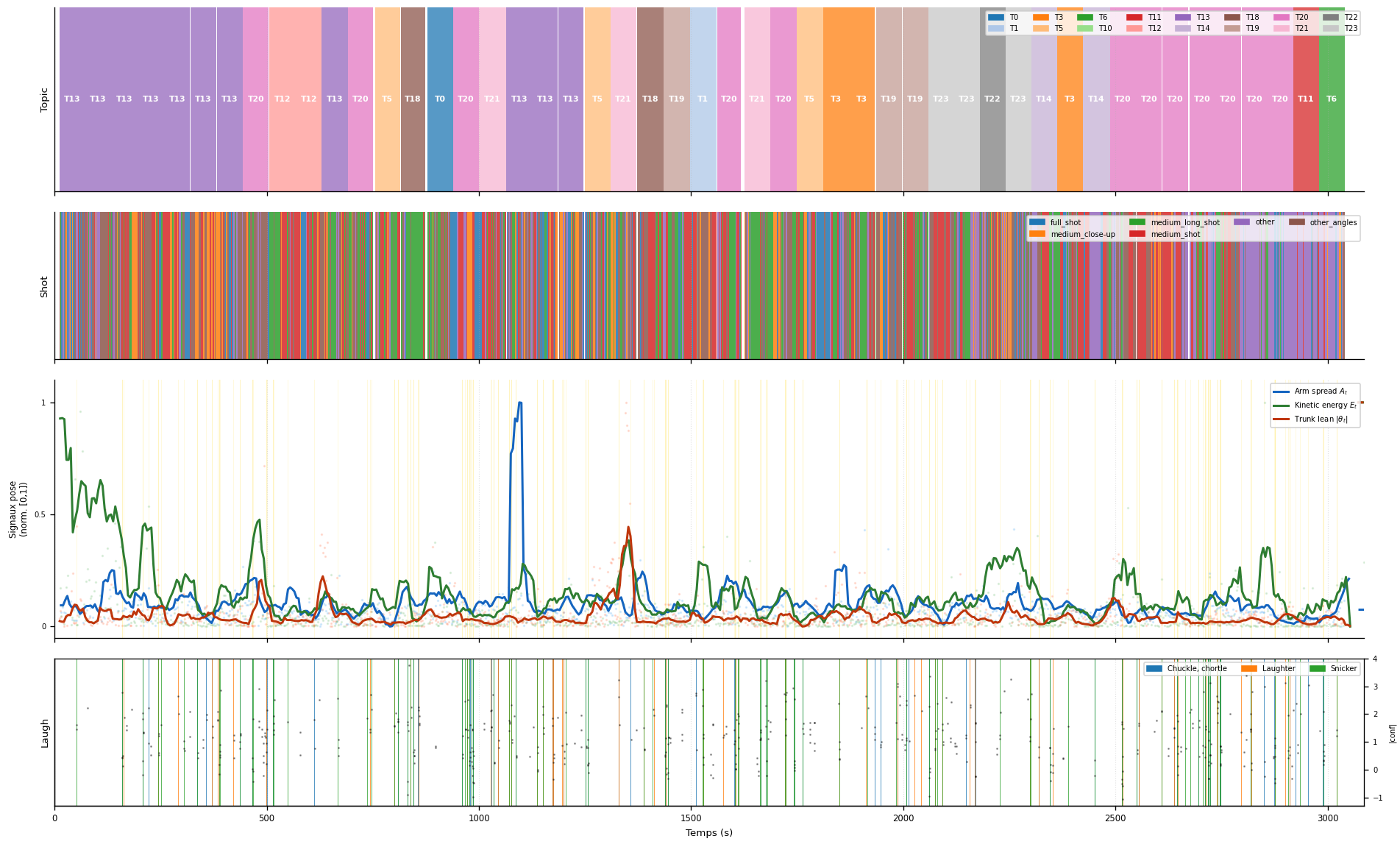}\\[2mm]
\caption{Example of aligned multimodal timelines for a show. Panels show from top to bottom: topic segments (BERTopic, 60\,s blocks), shot-type predictions (1\,Hz), three kinematic signals derived from raw pose keypoints (arm spread $A_t$, kinetic energy $E_t$, trunk lean $\theta_t$, each normalized to $[0,1]$; gold shading = laughter windows), and laughter events with confidence. The dense kinematic activity and prevalence of full-body shots reflect André's chaotic performance style; the alignment pipeline captures this without resampling across modalities.
}
\label{fig:viz_examples}
\end{figure*}

\subsection{Hierarchical Temporal Alignment}

Modalities operate on distinct temporal resolutions:  
\[
\Delta t_{\text{laugh}} = 0.8\,\text{s}, \quad
\Delta t_{\text{pose}} = 1\,\text{s}, \]
\[
\Delta t_{\text{shot}} = 1\,\text{s}, \quad
\Delta t_{\text{topic}} = 60\,\text{s}.
\]
To preserve native granularity, temporal alignment is performed through hierarchical containment rather than resampling.

Let each modality $m$ produce a sequence of temporal events or segments.
Topic segments serve as the \emph{anchor} level: each topic block $b_j = [s_j, e_j)$ defines a temporal window into which higher-frequency events are assigned by strict containment:
\[
e \text{ is assigned to } b_j \;\Longleftrightarrow\; t_e \in [s_j,\, e_j),
\]
where $t_e$ is the event timestamp. This applies uniformly to all nested streams: pose keyframes ($\Delta t{=}1$\,s), shot labels ($\Delta t{=}1$\,s), and laughter events ($\Delta t{=}0.8$\,s). Kinematic signals ($A_t$, $E_t$, $\theta_t$) are derived from the raw keypoints within each block after alignment.
Each topic segment also stores the 384-dimensional sentence-BERT embedding $\tilde{\mathbf{e}}_j$ computed during topic modeling (Section~\ref{sec:assignment}), enabling direct use for similarity retrieval or cross-show clustering without re-encoding.

\section{Corpus Output and Statistics}
\label{sec:corpus}

Following the pipeline in Section~\ref{sec:alignment}, we release only \emph{derived, time-aligned annotations} for 90 stand-up performances. No audio, image or video is distributed.

\subsection{Delivered outputs}
\begin{itemize}
    \item \textbf{Topics}: 60\,s segments with topic id, aggregated text, and sentence-BERT embedding.
    \item \textbf{Laughter}: contiguous events with start/end times, type label, and confidence score.
    \item \textbf{Shots}: one label per frame (1\,Hz) from YOLOv8-cls.
    \item \textbf{Poses}: raw 17-joint $(x,y)$ coordinates at 1\,fps with bounding-box dimensions; no prior clustering.
\end{itemize}

\begin{figure}[t]
\centering
\begin{minipage}{0.95\linewidth}
\begin{lstlisting}[language=json, caption={Excerpt from the unified dataset (V2 structure).}, label={fig:json}]
{
  "ID_1": {
    "metadata": {
      "show_id": "1",
      "n_blocks": 62,
      "embedding_dim": 384,
      "keypoint_joints": ["Nez", "Epaule_1", "...", "Cheville_2"]
    },
    "timeline": [{
      "block_id": 58,
      "start": 3480.0, "end": 3540.0,
      "topic_id": 6,
      "text": "marriage gender roles ...",
      "embedding": [0.021, -0.143, "..."],
      "laugh_events": [
        {"start": 3482.4, "end": 3485.6,
         "type": "laughter", "confidence": 0.92}
      ],
      "pose_keypoints": [{
        "time": 3483.0, "has_detection": true,
        "bbox": {"xmin": 412, "ymin": 28,
                 "xmax": 895, "ymax": 716},
        "keypoints": {
          "Epaule_1": [634.2, 182.5],
          "Poignet_1": [710.8, 480.3], "...": []
        }
      }],
      "shot_events": [
        {"time": 3483.0, "label": "full_shot",
         "class_id": 3, "score": 0.97}
      ]
    }]
  }
}
\end{lstlisting}
\end{minipage}
\end{figure}

\subsection{Summary statistics}
Average runtime per show is 63\,min (total $\approx$94\,h).  
The unified dataset contains 5,416 topic segments across 90 shows, each storing a 384-dim embedding.
The visual stream covers 322{,}973 frames at 1\,fps; 22\% are full-body frames yielding raw keypoint sequences for pose analysis.
Text yields $\approx$3{,}100 one-minute segments with non-outlier topic assignments.

\subsection{Cross-modal Analysis: Laughter Dynamics as a Use Case}
\label{sec:crossmodal}

We ask whether thematic content, kinematic profile, and shot composition co-vary with audience laughter across the 24 BERTopic topics and 5,416 aligned blocks.

\begin{figure}[h]
\centering
\includegraphics[width=\columnwidth]{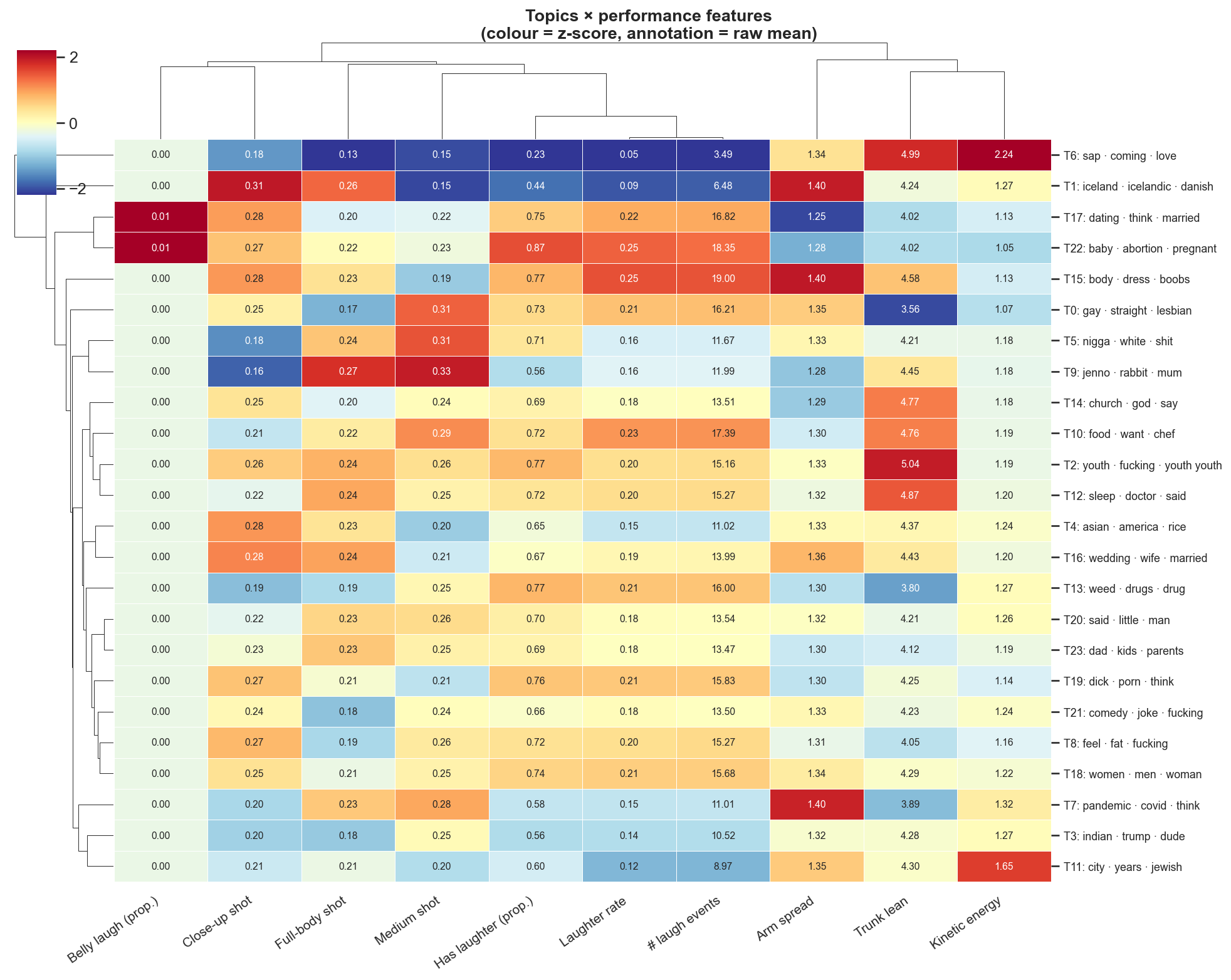}
\caption{Hierarchical clustermap of 24 BERTopic topics $\times$ 10 performance features (z-scored row-wise). Colour encodes standardized deviation from the per-feature mean. The dominant pattern separates a low-$\bar{E}_t$ / high-$\bar{r}_\ell$ cluster (top: T15, T22, T10 — personal/bodily topics) from a high-$\bar{E}_t$ / low-$\bar{r}_\ell$ cluster (bottom: T6, T11). T6 (sap/mae/hello) is a structural artefact corresponding to subtitle encoding markers and on-stage entry sequences; it should be excluded from content-level comparisons.}
\label{fig:clustermap_topics}
\end{figure}

\paragraph{Method.}
For each topic block, we compute (i) the laughter rate $r_\ell$, defined as the share of block duration covered by detected laughter events (mean coverage: 17.8\%, $\approx$1.2 events/10\,s); (ii) per-frame kinematic features (kinetic energy $E_t$, arm spread $A_t$, trunk lean $\theta_t$); and (iii) shot-type proportions (full-body, close-up, medium). These are averaged per topic, weighted by block count, to obtain topic-level profiles. Pearson correlations are then computed between each feature and the mean laughter rate $\bar{r}_\ell$ across the 24 topics. Figure~\ref{fig:clustermap_topics} presents the complete $24\,\text{topics} \times 10\,\text{features}$ matrix as a hierarchical clustermap; Table~\ref{tab:topic_laugh} reports the six most contrasted topics.

\begin{table}[h]
\centering
\small
\resizebox{\columnwidth}{!}{%
\begin{tabular}{p{3.6cm}rrrr}
\toprule
\textbf{Topic} & \textbf{n} & $\bar{r}_{\ell}$ & $\bar{E}_t$ & $\bar{A}_t$ \\
\midrule
T15: body / dress / boobs  &  97 & .253 & 1.13 & 1.40 \\
T22: baby / abortion       & 198 & .248 & 1.05 & 1.28 \\
T10: food / want / chef    & 210 & .230 & 1.19 & 1.30 \\
\midrule
T3: indian / trump / india & 197 & .141 & 1.27 & 1.32 \\
T1: iceland / icelandic    &  50 & .085 & 1.27 & 1.40 \\
T6$^\dagger$: sap / mae / hello &  81 & .051 & \textbf{2.24} & 1.34 \\
\bottomrule
\end{tabular}
}
\caption{Six most contrasted topics by mean laughter rate $\bar{r}_{\ell}$ (share of block duration covered by laughter events). $\bar{E}_t$: mean kinetic energy (normalized joint displacement between consecutive frames). $\bar{A}_t$: mean arm spread (wrist-to-wrist / shoulder-to-shoulder ratio). $^\dagger$T6 is a structural artefact; see Section~\ref{sec:discussion}.}
\label{tab:topic_laugh}
\end{table}

\paragraph{Finding 1 --- Kinetic energy negatively predicts laughter rate ($r=-0.75$).}
The strongest cross-modal signal across the 24 topics --- after the quasi-tautological event count --- is a \emph{negative} correlation between mean kinetic energy $\bar{E}_t$ and laughter rate: Pearson $r=-0.75$, $N=24$. Topics with the highest laughter rates all exhibit markedly low kinetic energy: T15 (body/dress, $\bar{E}_t=1.13$, $\bar{r}_\ell=0.253$), T22 (baby/abortion, $\bar{E}_t=1.05$, $\bar{r}_\ell=0.248$), T10 (food, $\bar{E}_t=1.19$, $\bar{r}_\ell=0.230$), all below the corpus mean of 1.31. The low-laughter end is anchored by T11 (city/tonight/jewish, $\bar{E}_t=1.65$, $\bar{r}_\ell=0.121$) and the artefactual T6 ($\bar{E}_t=2.24$, $\bar{r}_\ell=0.051$).

This pattern is consistent with a \emph{stillness-before-punchline} hypothesis: during high-laughter delivery, performers may reduce body movement and stabilize their posture to concentrate audience attention on the verbal content.

\paragraph{Finding 2 --- A thematic hierarchy of funniness.}
Topic content stratifies audience laughter systematically. The four topics with the highest laughter rates ($\bar{r}_\ell > 0.20$) are all personal and bodily in register: physical appearance (T15, $\bar{r}_\ell=0.253$), reproductive transgression (T22, $\bar{r}_\ell=0.248$; also the highest \emph{has\_laughter} rate across the corpus at 0.869), everyday life (T10, $\bar{r}_\ell=0.230$), and romantic relationships (T17, $\bar{r}_\ell=0.222$). By contrast, geopolitical and identity-framing topics generate substantially less laughter: T3 (trump/india, $\bar{r}_\ell=0.141$) and T1 (iceland, $\bar{r}_\ell=0.085$, 50 blocks concentrated in a single show). This replicates content-level funniness gradients documented in text-only humor classification \cite{yang-etal-2015-humor,annamoradnejad2024colbert}, anchoring them in live audience response at corpus scale.

\paragraph{Finding 3 --- Belly laughs are quasi-absent at topic granularity.}
The deepest laughter category (\emph{belly laugh}, AudioSet class 20) is effectively absent across all 24 topics: only T22 ($\hat{r}_{\text{belly}}=0.0051$) and T17 ($\hat{r}_{\text{belly}}=0.0061$) register non-zero values. Two non-exclusive explanations apply: (i) the Whisper-AT classifier may be conservative on this class (it represents fewer than 2 events per 74,000+ inference windows across the corpus); (ii) belly laughs are genuinely triggered by specific delivery moments rather than by sustained thematic content, making them invisible at 60\,s block granularity. Either interpretation suggests that capturing deep, distinctive laughter requires event-level annotation at a finer temporal resolution.

\paragraph{Finding 4 --- Shot composition and reactive montage ($r=+0.28$).}
Close-up shot proportion shows a weak positive correlation with laughter rate ($r=+0.28$, $N=24$). High-laughter topics T15 and T22 both display above-average close-up ratios (0.278 and 0.265 respectively), consistent with \emph{reactive montage}: directors increase close-up coverage during high-laughter passages, foregrounding the performer's facial expression at punchline delivery. A notable exception is T1 (iceland/icelandic), which exhibits the highest close-up ratio in the corpus (0.315) despite a low laughter rate ($\bar{r}_\ell=0.085$), indicating that this association is partially confounded by performer- and show-level filming conventions.

\subsection{Short-horizon Laughter Onset Prediction}
\label{sec:onset}

Given the multimodal stream up to time $t$, can we predict whether a new laughter event will begin in the next $\delta=2$\,s?

\paragraph{Task and experimental setup.}
For each show, we sample anchor points at 1\,s steps, excluding moments inside an ongoing laughter event. This yields 285,916 anchors across 90 shows; the positive rate is 17.0\% (a new onset occurs in the next 2\,s). Shows are split at the group level (GroupShuffleSplit): 62 shows for training, 14 for validation, 14 for test---no show appears in more than one split. A HistGradientBoostingClassifier is trained with balanced sample weights; the decision threshold is tuned on validation by maximizing F1.

Three feature groups are defined. \textbf{History} (10 scalars): from laughter events in the past 10\,s---event count, rate, coverage, maximum event duration, mean and maximum confidence, coverage in the last 2\,s and 5\,s, time since last onset, time since last end. \textbf{Text} (64 scalars): the current topic block's 384-dim sentence-BERT embedding, reduced to 64 dimensions via PCA fitted on the training set only. \textbf{Vision} (20 scalars): shot proportion histogram over 6 classes, shot change rate, mean shot confidence, and 12 pose scalars (arm spread mean/std/max/trend, trunk lean mean/std, kinetic energy mean/std/max/trend, detection rate)---all aggregated over the past 10\,s window.

\begin{table}[h]
\centering
\small
\resizebox{\columnwidth}{!}{
\begin{tabular}{lrrrrr}
\toprule
\textbf{System} & \textbf{AUROC} & \textbf{AUPRC} & \textbf{F1} & \textbf{Prec} & \textbf{Rec} \\
\midrule
history-only            & 0.643 & 0.275 & 0.336 & 0.223 & 0.682 \\
text-only               & 0.554 & 0.197 & 0.297 & 0.177 & 0.926 \\
vision-only             & 0.538 & 0.187 & 0.291 & 0.178 & 0.806 \\
text + vision           & 0.577 & 0.210 & 0.300 & 0.182 & 0.867 \\
text + vision + history & \textbf{0.647} & \textbf{0.277} & \textbf{0.342} & \textbf{0.248} & 0.553 \\
\midrule
Random (AUPRC baseline) & --- & 0.170 & --- & --- & --- \\
\bottomrule
\end{tabular}
}
\caption{Short-horizon laughter onset prediction: ablation over feature groups. Positive rate = 0.170. Test set: 45,894 anchors from 14 held-out shows. AUPRC of a random classifier equals the positive rate. Threshold tuned on validation for F1/precision/recall.}
\label{tab:onset}
\end{table}

\paragraph{Findings.}
First, temporal laughter history is by far the strongest individual predictor (AUROC\,=\,0.643), leaving only marginal room for the other modalities: the best multimodal system (text+vision+history) improves AUROC by only 0.004 over history alone. This reflects a \emph{temporal auto-correlation} of audience laughter: a hot room stays hot, independently of what is being said or shown. Second, vision-only is the weakest unimodal system (0.538), but combining it with text (0.577) outperforms both text-only (0.554) and vision-only---a consistent, if small, multimodal synergy. Third, adding text and vision to history improves precision from 0.223 to 0.248 while moderately reducing recall---the full model is less trigger-happy, reducing false positives. Fourth, the overall performance level (AUPRC\,=\,0.277 vs.~random 0.170, a $1.6\times$ lift) indicates that laughter onset is predictable above chance from a 10\,s window, but far from deterministic: the stochastic nature of audience response and the 60\,s temporal granularity of topical context both limit the ceiling of short-horizon prediction.

\subsection{Data Structure and Access}

Annotations are serialized as a hierarchical JSON per show (Figure~\ref{fig:json}); each topic block stores its four aligned streams, enabling direct temporal queries without resampling.

\subsection{Multimodal Visualization Examples}
\label{sec:visualizations}

Figure~\ref{fig:viz_examples} illustrates the aligned multimodal timeline for one show; analogous visualizations for all 90 specials are distributed with the corpus as an interpretability and consistency check for the alignment pipeline.

\section{Discussion}
\label{sec:discussion}

The descriptive use case (Section~\ref{sec:crossmodal}) yields four findings. (1) The $E_t$--laughter anti-correlation ($r=-0.75$) is consistent with a \emph{stillness-before-punchline} pattern, but remains correlational: filming conventions, performer mobility, and the artefactual T6 are plausible confounders; event-level replication (kinetic energy in the 5\,s before vs.\ after laughter onset) would be a stronger test. (2) Personal/bodily topics outperform geopolitical ones on laughter rate, replicating text-only funniness gradients at the level of live audience response and suggesting thematic content is a first-order predictor independently of delivery style. (3) The near-absence of belly laughs at 60\,s granularity motivates finer annotation: deep laughter is likely tied to specific delivery moments invisible at block level. (4) The shot composition signal ($r=+0.28$), coherent with reactive montage, is partially confounded by show-level filming conventions.

The predictive use case (Section~\ref{sec:onset}) adds three observations. (5) Laughter auto-correlation (history-only AUROC\,=\,0.643) accounts for most predictable variance, consistent with crowd contagion~\cite{provine1992contagious}. (6) Text and vision contribute marginally to precision (0.248 vs.\ 0.223), limited by 60\,s block granularity; sentence-level and frame-level features would likely yield stronger gains. (7) The modest ceiling (AUPRC\,=\,0.277 vs.\ 0.170 random) reflects the inherent stochasticity of audience response, absent from our single-recording specials.

Together, these results show that hierarchical temporal alignment enables both descriptive and predictive cross-modal analyses while exposing their respective limits. The resource's value lies in enabling comparable, interpretable measurements of multimodal synchrony at corpus scale rather than absolute claims about funniness. Because all shows are professionally edited Netflix specials, platform conventions are embedded in the signal; comparisons are most reliable within similarly produced shows. Next steps include event-level annotation, prosodic features, and intra-comedian analyses to disentangle performer style from content.


\section{Conclusion}

This lightweight and modular pipeline provides an effective, reusable, and scalable framework for modeling the multimodal structure of stand-up comedy — a form whose apparent simplicity belies the artistry and craftsmanship of its performers. By operationalizing core dimensions such as gesture, timing, and audience response — a conceptual and technical challenge — and thus offering a model of stand-up performance, it supports both systematic analysis and critical reflection on what remains beyond computation.

\paragraph{Future work.} While our pipeline was applied across a wide variety of performers and stand-up comedy specials, future experiments could focus on multiple video recordings by the same comedian, in order to track stylistic evolution over time and test the modularity of performance elements — a relevant hypothesis for stand-up, where long-form shows are often assembled from recombined short routines.  In our current setup, editing is treated as part of the artistic object — and while some editing conventions recur, their variability across shows may obscure key aspects of the live performance. Obtaining recordings from other sources would help generalization from this database. Capturing original data would also open new avenues toward the study of more local, ephemeral, or amateur practices, anchored in specific socio-geographic contexts.

\section{Code Availability}
All code used for processing and analysis is available at the following anonymous repository: \url{https://github.com/depotanonyme/16102025}.

\section{Acknowledgements}

 We are thankful to participants at the AV in DH workshop (George Mason University), at the Sciences Po Medialab seminar and at the Bridging Computational Humanities and Computational Social Science Workshop (Ecole nationale des chartes - PSL) for their insightful remarks. We particularly thank Sylvaine Guiot and Jean-Philippe Cointet for their guidance. Errors remain our own

\section{Bibliographical References}\label{sec:reference}

\bibliography{lrec2026-example}

\end{document}